\documentclass{article} 
\usepackage{collas2024_conference,times}
\usepackage{easyReview}
\usepackage{subcaption}
\usepackage{wrapfig}
\usepackage{amsmath}
\usepackage{bbm}

\usepackage{amsmath,amsfonts,bm}

\def\eqref#1{equation~\ref{#1}}

\def\1{\bm{1}}

\DeclareMathAlphabet{\mathsfit}{\encodingdefault}{\sfdefault}{m}{sl}
\SetMathAlphabet{\mathsfit}{bold}{\encodingdefault}{\sfdefault}{bx}{n}

\DeclareMathOperator*{\argmax}{arg\,max}

\usepackage{hyperref}
\hypersetup{
    colorlinks=true,
    linkcolor=red,
    filecolor=magenta,
    urlcolor=blue,
    citecolor=purple,
    pdftitle={Overleaf Example},
    pdfpagemode=FullScreen,
    }

\title{Patch-Based Contrastive Learning and Memory Consolidation for Online Unsupervised Continual Learning}

\author{Cameron Taylor\textsuperscript{1}\\
{\tt\small cameron.taylor@gatech.edu} 
\and
\bf Vassilis Vassiliades\textsuperscript{2}\\
{\tt\small v.vassiliades@cyens.org.cy} 
\and
\bf Constantine Dovrolis \textsuperscript{1,3}\\
{\tt\small constantine@gatech.edu} 
\And
\normalfont{\textsuperscript{1}Georgia Institute of Technology} \\
Atlanta, GA
\and
\textsuperscript{2}CYENS Centre of Excellence \\
Nicosia, Cyprus
\and
\textsuperscript{3}The Cyprus Institute \\
Nicosia, Cyprus
}

\collasfinalcopy 

\begin{document}

\maketitle

\begin{abstract}

    We focus on a relatively unexplored learning paradigm known as {\em Online Unsupervised Continual Learning} (O-UCL), where an agent receives a non-stationary, unlabeled data stream and progressively learns to identify an increasing number of classes. This paradigm is designed to model real-world applications where encountering novelty is the norm, such as exploring a terrain with several unknown and time-varying entities. Unlike prior work in unsupervised, continual, or online learning, O-UCL combines all three areas into a single challenging and realistic learning paradigm. In this setting, agents are frequently evaluated and must aim to maintain the best possible representation at any point of the data stream, rather than at the end of pre-specified offline tasks. The proposed approach, called \textbf{P}atch-based \textbf{C}ontrastive learning and \textbf{M}emory \textbf{C}onsolidation (PCMC), builds a compositional understanding of data by identifying and clustering patch-level features. Embeddings for these patch-level features are extracted with an encoder trained via patch-based contrastive learning. PCMC incorporates new data into its distribution while avoiding catastrophic forgetting, and it consolidates memory examples during  ``sleep" periods. We evaluate PCMC's performance on  streams created from the ImageNet and Places365 datasets. Additionally, we explore various versions of the PCMC algorithm and compare its performance against several existing methods and simple baselines. The code is publicly available \href{https://github.com/CameronTaylorFL/upl-benchmark}{on Github}. 
\end{abstract}

\section{Introduction}
Imagine an agent navigating an unfamiliar environment with objects that have never appeared in its  pre-training data. As the agent explores that environment, it must recognize new classes of objects. Furthermore, it must generalize from the observed data to other instances of the same class. Additionally, it should not forget previously learned classes, even if it does not encounter such instances often. This setup demands an efficient learning model where the agent operates in a streaming fashion, retaining only a minuscule fraction of the observed data due to storage or privacy constraints. We refer to this learning paradigm as  {\em Online Unsupervised Continual Learning} or ``O-UCL'' for short.

The O-UCL learning paradigm must address the following three challenges: 1) The data stream is non-stationary, in the sense that the number of classes of objects in the stream increases with time; 2) the observed data is unlabeled, and so the agent needs to identify novel classes in real-time and without supervision; 3) the agent cannot store the observed data for future replay, necessitating  online stream learning. While a simple solution might be to utilize a frozen encoder pretrained on the largest possible dataset, this is insufficient for applications in which the environment is inherently unknown and/or it includes novelty (e.g., people, animals or objects that were never previously seen). Instead, we consider a dynamic approach in which the encoder and the corresponding data representations are periodically adapted during short ``sleep'' periods.

\begin{figure*}[t!]
    \centering
    \includegraphics[width=0.99\textwidth]{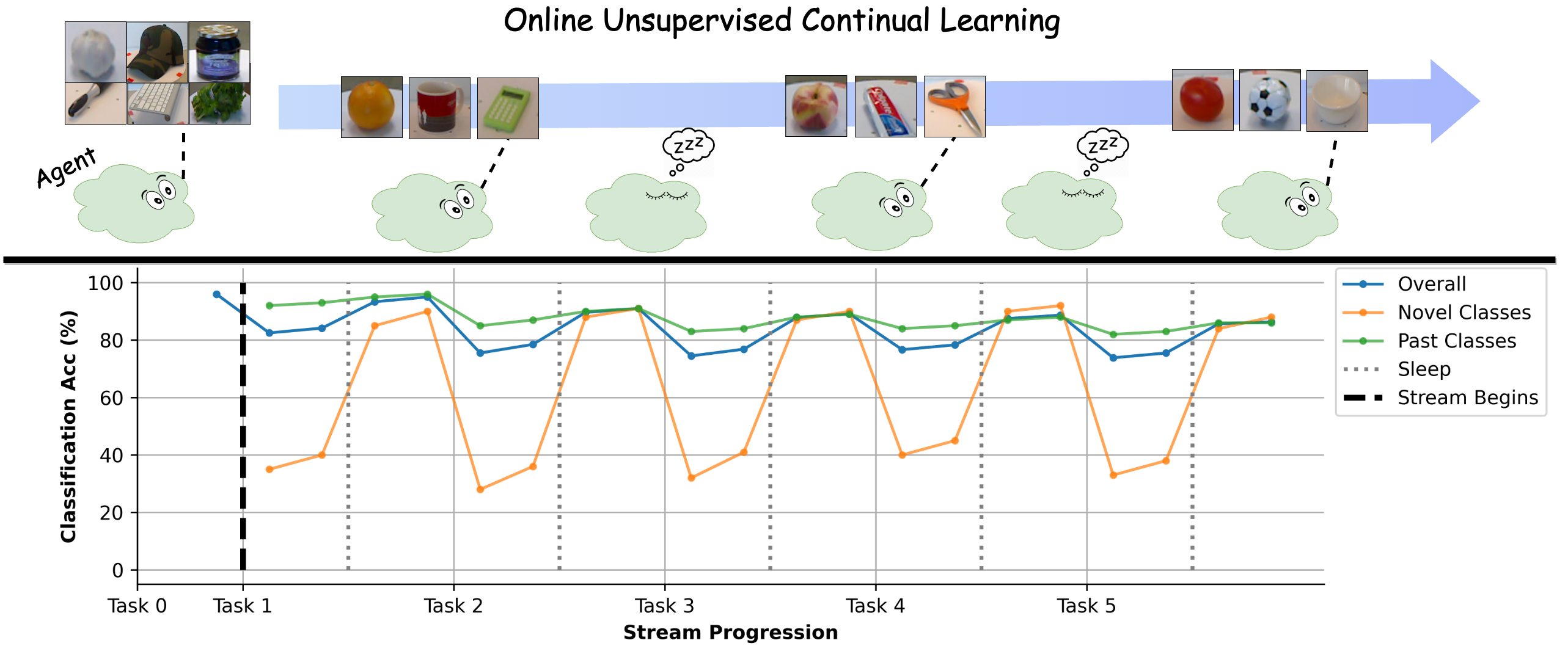}
    \caption{A toy example for an O-UCL scenario. After an offline initialization (task-0), the agent is presented with a stream of data consisting of images from various classes (say three new classes in each task). The agent is tasked with learning to identify both new and previously known classes, without forgetting classes that no longer appear in the stream. The performance is evaluated frequently during the stream, to monitor how the agent learns over time. During sleep periods, the agent retrains its encoder and adapts its stored  representations. Note that a small number of labeled examples are given to the classifier only during inference -- no labeled examples are available for representation learning during the stream.}
    \vspace{-1em}
    \label{fig:cornerstone}
\end{figure*}

For example consider the following hypothetical applications: an AI-powered  drone that explores a new construction site or physical system, or a face recognition system that has to classify people it has never seen before with distinct identifiers. In addition to the three O-UCL challenges described above, such applications have one more common requirement: there is no boundary between training and inference. Instead, a successful O-UCL learner should both learn and perform its task during the data stream, exhibiting gradually better performance as it observes more data.
In Figure \ref{fig:cornerstone}, we demonstrate a toy scenario to exemplify O-UCL. 

A similar problem to O-UCL was first introduced in~\citep{smith2021unsupervised}, referred to as ``Unsupervised Progressive Learning.'' That work also proposed a method called ``Self-Taught Associative Memories'' or STAM. The method we propose here, referred to as \textbf{P}atch-based \textbf{C}ontrastive learning and \textbf{M}emory \textbf{C}onsolidation or ``PCMC,'' utilizes some techniques from STAM but it also introduces several new ideas. PCMC operates in a cycle of ``wake" and ``sleep" periods. The model identifies and clusters incoming stream data during each wake period, while it retrains the encoder and consolidates data representations during each sleep period. PCMC utilizes a novel patch-based contrastive learning encoder along with online clustering and novelty detection techniques. We evaluate PCMC's performance against several baselines on challenging natural image datasets such as ImageNet~\citep{deng2009imagenet} and Places365~\citep{zhou2017places}.

\section{PCMC Method}
\label{method:PCMC}
The key component of PCMC  is a growing set of cluster centroids that represent distinct data features. To learn useful centroids,  we utilize an encoder $F_{\phi}(\cdot)$, which is a deep neural network with parameters $\phi$ that generates similar embeddings for similar inputs and dissimilar embeddings for dissimilar inputs (contrastive learning). As the distribution of the stream changes over time, the encoder must adapt to changes in the data distribution (e.g., novel classes) but also to avoid concept drift and catastrophic forgetting. PCMC accomplishes this by adopting a wake-sleep cycle. During a wake period, the model performs novelty detection (i.e., creation of new clusters) and it also adapts the centroids of previously known clusters with new data. During a sleep period, the encoder is retrained and also the  representations of the learned centroids are updated.

Critically, instead of clustering entire input examples (images in this paper), PCMC breaks each incoming example into smaller patches, and maps each patch to a cluster that represents a distinct feature of the data. The use of patches is important for two reasons. The first is that as the data distribution changes, we want to leverage potential forward transfer for shared features across classes (e.g., fire truck tires and ambulance tires) at the level of patches. Secondly, clustering similar inputs is more challenging at the level of the entire input compared to the patch level. This is because the variance of a patch-level feature (e.g., a dog's nose) is typically much smaller than the variance of the entire input (e.g., an image of a dog). By learning features at the patch level, the model can build a compositional understanding of each class, based on the class-informative features it includes. 

\begin{figure*}[t!]
    \centering
    \includegraphics[width=0.99\textwidth]{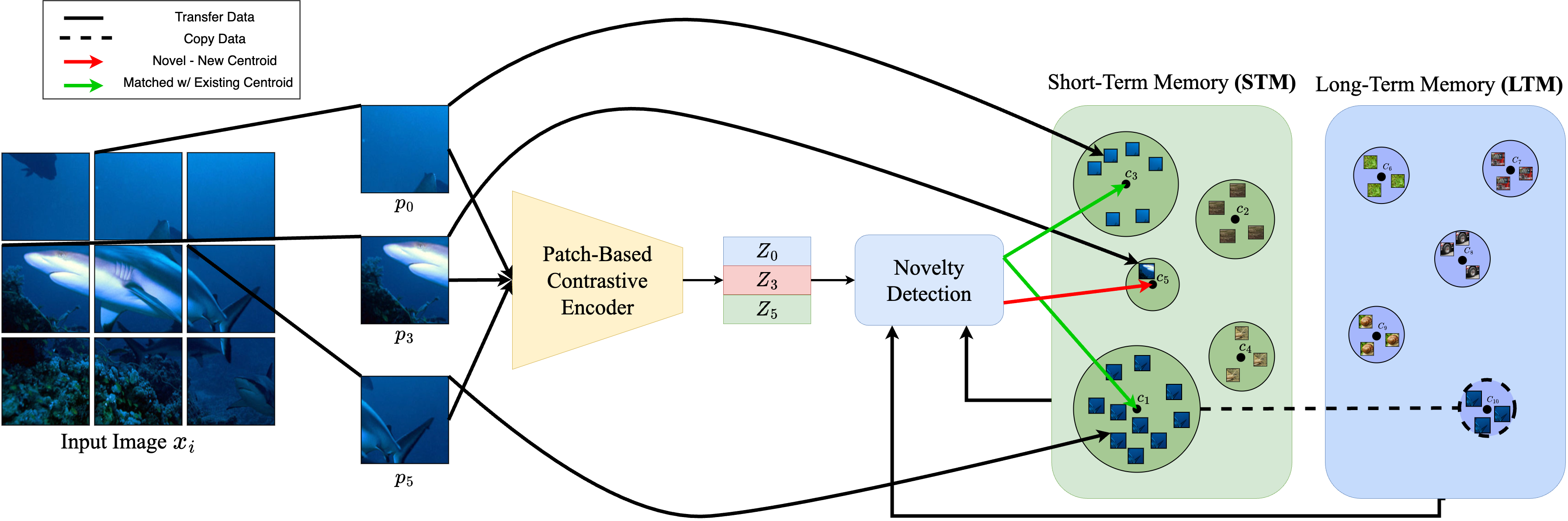}
    \caption{This figure summarizes the wake period of PCMC. The input $x_i$ is broken up into patches, and the encoder $F_{\phi_{s}}$ generates an embedding for each patch. A patch embedding is compared with existing centroids to perform novelty detection. If the embedding is far from any stored centroid, a new cluster is created in Short-Term Memory. Otherwise, that patch is mapped to its nearest centroid -- and the location of the latter is updated. When a cluster accumulates several ($\theta$) patches, it is copied from Short-Term Memory to Long-Term Memory so that it is never forgotten.}
    \vspace{-1em}
    \label{fig:centroid-learing}
\end{figure*}

\subsection{Patch-Based Contrastive Learning}
\label{method:patch-contrastive}
To learn an encoder with the previous properties, we propose a patch-based modification of typical instance-based contrastive learning. Given an input image $x$, we extract patches $\mathcal{P} = \{{\bf p_i}\}_{i=1}^{N}$ and apply a set of augmentations (see Section~\ref{sec:experiments}) to each patch to generate $\bf p_{i,1}$ and $\bf p_{i,2}$. To maximize the number of training samples extracted from each image, we extract highly overlapping patches using a stride that is smaller than the patch size. We ensure that highly overlapping patches are always in separate batches. We also experimented with the SupCon loss from~\citep{khosla2020supervised}, where any highly overlapping patches would be given the same label. However, we observed better results when simply splitting  overlapping patches into different batches. In this work, we utilize the loss from~\citep{chen2020simple}, but our framework is compatible with any instance-based contrastive learning loss. The loss function for a single positive pair of patch embeddings $(\bf z_i, \bf z_j)$ is given by:
\begin{equation}
    \mathcal{L} = -\text{log} \frac{\text{exp}(\text{sim}(\bf z_i, \bf z_j) / \tau)}{\sum_{k=1}^{2N} \mathbbm{1}_{[k \neq i]}\; \text{exp}(\text{sim}(\bf z_i, \bf z_k) / \tau)}
\end{equation}
where $N$ is the total number of patches in the batch and sim($\cdot, \cdot$) is the cosine similarity.

\subsection{Wake Phase}
\label{method:wake}

\paragraph{Centroid Learning:}
\label{method:cent-learn}
When a new input $\bf x$ is received, the model extracts patches of $\bf x$ based on a given patch size and stride. These patches are then encoded with  
$F_{\phi}$. The patch embeddings are then processed sequentially using an online clustering algorithm: given a patch $\bf p$, we find the most similar cluster by mapping the embedding of that patch  $\bf z$ with the nearest cluster centroid in the set of existing centroids $\mathcal{C}$ as
\begin{equation}
    {\bf c_{j}} = \arg\min_{\bf{c} \in \mathcal{C}} d({\bf z,{\bf c})}
\end{equation}
where $d(\cdot, \cdot)$ can be any distance metric;  we use cosine distance. If the nearest cluster is within the \textit{novelty detection} threshold (see later in this section), we expect that the cluster consists of patches that have high similarity with $\bf p$ and assign $\bf p$ to cluster $\bf c_j$. When a patch is assigned to a cluster, its embedding updates the centroid as follows, 
\begin{equation}
{\bf c_{j}} = (1 - \alpha) {\bf c_{j}} + \alpha \, {\bf z} 
\label{eq:centroid_update}
\end{equation}
and the raw patch $\bf p$ is  temporarily stored with the selected centroid in an auxiliary memory $\mathcal{M}_j$. The centroid learning rate $\alpha$ controls how much influence an individual input has on the centroid, and it can be set as a fixed value or it can decay as the size of the cluster increases; we do the former.

\paragraph{Novelty Detection:}
\label{method:novelty}
If the distance between $\bf z$ and its nearest cluster centroid $\bf c_j$ is greater than the novelty detection threshold ${\bf \hat{D}}$, a new cluster is created and ${\bf z}$ becomes the centroid of the new cluster. The raw patch ${\bf p}$ is also temporarily stored in the new cluster's memory $M$. Because the data distribution changes over time, the threshold ${\bf \hat{D}}$ should also be dynamically updated. We estimate ${\bf \hat{D}}$ in an online manner by maintaining a distance distribution between recent patch embeddings ${\bf z}$ and their nearest centroid ${\bf c_j}$, using a sliding window over recently observed samples. The novelty detection threshold is defined as a high percentile ($\beta$) of this distance distribution.  

\begin{wrapfigure}[23]{r}{0.55\textwidth}
    \centering
    \includegraphics[width=0.54\textwidth]{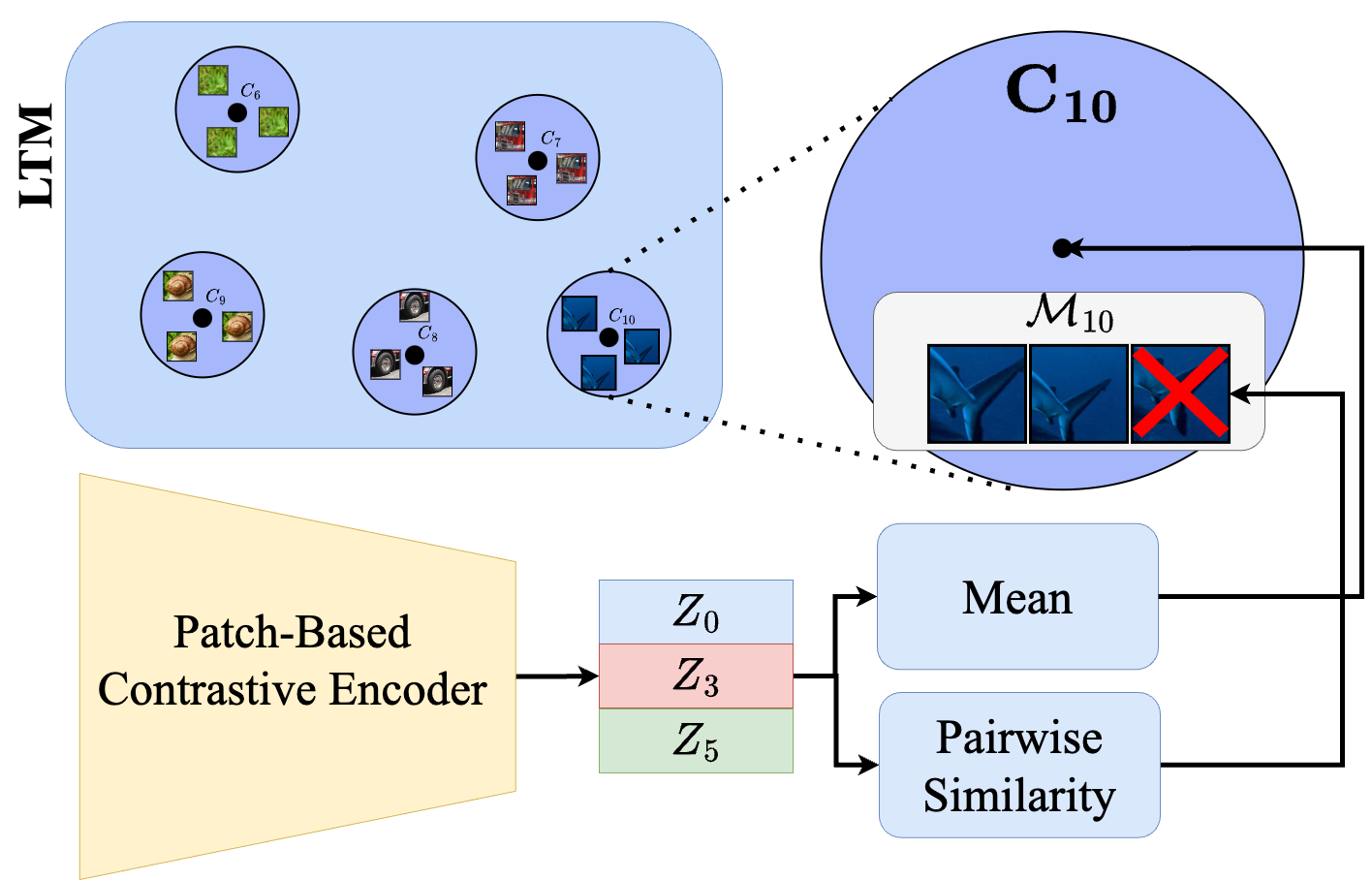}
    \caption{The memory consolidation process during the sleep phase of the PCMC algorithm. Each centroid in the model's LTM is recomputed using the updated contrastive encoder. Very similar examples stored in the centroid's memory are pruned.}
    \vspace{-1em}
    \label{fig:sleep-diagram}
\end{wrapfigure}

\paragraph{Short \& Long Term Memory:}
\label{method:dual-memory}
The model partitions the learned cluster centroids into two groups: the short-term memory (\textbf{STM}) and the long-term memory (\textbf{LTM}). The STM maintains  centroids that were recently initialized. It has a fixed size and uses a least-recently used replacement policy to create space for new centroids, once it has reached capacity. The purpose of the STM is to temporarily store clusters until they have been matched multiple times ($\theta$), allowing the model to be more certain that this cluster represents an actual feature of the data, and not  an outlier or noise. Whenever a centroid $\bf c_j$ in the STM has reached $\theta$ matches, it is copied to the LTM  along with its  memory $\mathcal{M}_j$. The STM centroid $\bf c_j$ is marked as unusable until it is evicted or the model goes to sleep (more details in~\ref{method:sleep}). The purpose of the LTM is to permanently store useful features that have been observed multiple times in the stream. Those cluster centroids are frozen to prevent concept drift and catastrophic forgetting, while the corresponding cluster memory is utilized to retrain the encoder during sleep periods, as explained next.

\subsection{Sleep Phase}
\label{method:sleep}
Periodically, PCMC goes to sleep to update the encoder based on the recently seen  data since the last sleep period. Additionally, sleep is used to consolidate the training patches stored in LTM. The sleep phase consists of two stages. The first is the encoder retraining stage, which utilizes the stored patch examples to fine-tune the encoder's weights. The second is the centroid update stage, which updates the centroids' positions and stored examples, avoiding concept drift and improving efficiency. 

\paragraph{Encoder Retraining:}
\label{method:encoder-update}
PCMC creates a new dataset from the set of patch examples stored in STM and LTM, which we call $X_{\text{sleep}}$. The STM data is meant to provide new information from the most recently created clusters, while the LTM data serves to remember previously learned centroids to avoid catastrophic forgetting. The encoder $F_{\phi_{s-1}}(\cdot)$, with weights $\phi_{s-1}$ after $s-1$ sleep cycles, is then updated using the contrastive learning method of Section~\ref{method:patch-contrastive}. 

\paragraph{Centroid Updates:}
\label{method:centroid-update}
After the encoder is updated, PCMC updates the centroids in both STM and LTM to avoid suffering from concept drift. The model re-embeds each centroid using $K$ examples from that centroid's memory $\mathcal{M}_j$ as
\begin{equation}
    {\bf c_{j}} = \frac{1}{K}\sum_{i=1}^{K} F_{\phi}(\mathcal{M}_{j, i})
\end{equation}
In the experiments of this paper, PCMC utilizes a single example ($K = 1$) as opposed to the average of all stored examples. The single example is chosen as the first example added, which is also the example used to originally create the centroid. After that point, the STM example memories $M$ are erased, and the counter of STM matches is reset.

\paragraph{Memory Consolidation:}
\label{method:mem-consolidate}
Once the centroid positions have been updated, PCMC utilizes a targeted forgetting strategy to help prune some of the more redundant examples stored in LTM. At this stage, each centroid $\bf c_j$ is assigned a probability $P_{\text{prune}}$ of pruning a single example from its own memory examples $\mathcal{M}_j$. Specifically, the probability scales with the number of examples  stored and is given by
\begin{equation}
\label{eq:mem-prune}
    P_{\text{prune}}(\mathcal{M}_j) = \left( \frac{|\mathcal{M}_j| - M_{\text{min}}}{M - M_{\text{min}}} \right)^{k}
\end{equation}
where $|\mathcal{M}_j|$ is the number of examples  stored with the $j$'th centroid, $M_{min}$ is a hyper-parameter controlling the minimum number of examples a centroid can have, and $k$ is a scaling factor that impacts the frequency of pruning. This equation allows us to select a pruning probability that decreases from $1$ to $0$ based on a desired maximum ($M$) and minimum ($M_{\text{min}}$) number of stored examples, while $k$ controls how aggressive the pruning is as $\mathcal{M}_j$ gets closer to $M_{\text{min}}$. In Section~\ref{sec:mem-perf} we showcase how changing $M$, or not performing this consolidation step, impacts performance and also the overall memory requirements of the PCMC model. Further experiments with $k$ and $M_{\text{min}}$ are given in the Appendix.

\subsection{Initialization}
\label{method:init}
To bootstrap the initial novelty detection values and encoder weights, the model starts with an initialization task $T_0$ that consists of unlabeled images. The data $X(T_0)$ is available upfront. The model trains on the data using the approach described in \ref{method:patch-contrastive}. After training, we utilize k-means to cluster a randomly sampled subset of $X(T_0)$ into $\Delta_0$ clusters, and initialize both  STM and LTM with these centroids. In the LTM, each centroid also stores $M_{init}$ raw patches that are members of that centroid's cluster so they can be used later during the sleep phases. We also initialize $\hat{D}$ by sampling a small set of distances between random patches from $X(T_0)$ and the nearest centroid in the initialized STM. The initial novelty detection threshold $\hat{D}$ is set to the $\beta$'th percentile of that distance distribution

\section{Experimental Setup}
\label{sec:experiments}

In this section, we describe the datasets, training details, and evaluation methods used in our experiments.

\subsection{Datasets and Streams}
We create streams from two different datasets: ImageNet~\cite{deng2009imagenet} and Places365~\cite{zhou2017places}. The ImageNet-derived stream focuses on object classification, while the Places365 stream focuses on scene classification. Both streams consist of 40 classes and are referred to as ImageNet-40 and Places365-40, respectively. Each stream comprises 16 tasks: the first task (T0) is offline and contains 10 classes, while the remaining 15 tasks each contain two classes, forming the stream. We resize each image to $120 \times 120$ pixels and normalize them. Tables containing all hyperparameters for PCMC and baselines can be found in the Appendix.

\subsection{Training Details}

\paragraph{Initialization:}
We use a ResNet18~\cite{he2016deep} as the encoder backbone and a two-layer MLP with a hidden size of 512 for the projection head during contrastive training. The encoder is trained for 500 epochs using the SGD optimizer with an initial learning rate of 0.6, a weight decay of 1e-5, a momentum of 0.9, and a cosine annealing learning rate schedule. For the initial novelty detection threshold estimate and initial STM/LTM centroids, we utilize a subset of the T0 data as described in Section~\ref{method:init}.

\paragraph{Sleep Cycle:}
In our primary experiments, PCMC goes to sleep at fixed intervals during each task. In \ref{res:sleep-timing}, we consider the frequency and timing of sleep cycles, demonstrating that timing is not critical to overall performance. During each sleep cycle, we train the ResNet18 backbone for 300 epochs using the same optimization hyperparameters as during initialization. Since the sleep dataset stores patches without reference to the original image, each training batch incontrastive learning consists of independent patches, not entire images.

\paragraph{Augmentations:}
We use a similar set of augmentations to~\cite{chen2020simple} -- but with some tweaks to optimize for our patch-based approach. We reduce the magnitude of \textit{\bf ColorJitter} and remove \textit{\bf GaussianBlur} entirely to emphasize color and texture in defining visual similarity. Additionally, we modify \textit{\bf RandomCropandResize} by applying the crop only to the first augmented version, while keeping the range of possible crop sizes for the second version closer to the full size. This encourages invariance to small changes in translation and scale, while avoiding overly drastic transformations.

\subsection{Evaluation}
For classification and clustering tasks, we use an approach similar to~\cite{smith2021unsupervised}, as explained next.

\paragraph{Classification with PCMC:}
To perform classification, PCMC takes a small set of labeled images $X_L$ and breaks each image into patches. For a patch $p$ from an image of class $k$, each centroid gets an association distribution $g_{\bf c}$ based on
\begin{equation}
g_{\bf c}(k) = \frac{\sum_{{\bf p} \in X_L(k)}{D({\bf p, c_j)}}} {\sum_{k' \in L(t)}\sum_{{\bf p} \in X_L(k')}{D({\bf p,c_j})}}, \quad k=1 \dots L(t),
\end{equation}
where $D(\bf p, c_j)$ is the distance between patch $p$ and its nearest centroid $c_j$. Centroids that are strongly associated with one class are considered ``class informative". PCMC then breaks down test images $X_T$ into patches, finds the most similar centroid $c_j$ for each patch, and if that centroid is class informative, it becomes an eligible voter. Eligible voters cast a vote for class $k^*$ based on 
\begin{equation}
 k^* = \argmax_{k \in L(t)} g_{\bf c_j}(k)   
\end{equation}
with the vote's magnitude equal to $g_{\bf c_j}(k^*)$. The total vote for class $k$ is the sum of votes for $k$, and the final class prediction is the highest voted class.

\paragraph{Clustering with PCMC:}
\label{app:clustering}
To perform tasks that do not require any labels, such as offline clustering, we use PCMC features. For each patch of input $\bf{x}$, we compute the nearest LTM centroid. The set of all such centroids across all patches of $\bf{x}$ is denoted by $\Phi(\bf{x})$. Given two inputs $\bf{x}$ and $\bf{y}$, their pairwise distance is the Jaccard distance of $\Phi(\bf{x})$ and $\Phi(\bf{y})$. 

To perform clustering on a given set of inputs and a target number of clusters, we apply a standard spectral clustering algorithm on all pairwise distances $\Phi(\bf{x})$ and $\Phi(\bf{y})$. Other clustering algorithms can also be used.

\subsection{Baselines}

\paragraph{Whole-Image Baseline:}
This baseline exemplifies the wake/sleep cycle used with PCMC, but without the patch-based approach. It uses contrastive learning and it stores entire images (instead of patches) over time. Specifically, the model randomly selects and stores $M_{init}$ images from T0 and $M$ images from each task, adding them to a long-term memory (LTM). It periodically goes to sleep and retrains the encoder using a contrastive objective on the LTM images. The $M$ and $M_{init}$ values are slightly more than the actual memory usage of PCMC to ensure a fair comparison. The whole-image baseline trains for 500 epochs during T0 and 300 epochs for each subsequent sleep cycle. The encoder backbone is a ResNet18~\cite{he2016deep} with a two-layer MLP for the projection head, using the SimCLR loss and augmentations.

\paragraph{SCALE \cite{yu2022scale}}:
We modify SCALE to include pretraining during T0, using its contrastive loss. The memory bank size is fixed, equivalent to the final memory utilization of PCMC. We use the same hyperparameters with SCALE for TinyImageNet~\cite{le2015tiny}, with slight modifications to batch size and learning rate for the O-UCL setting. We also explored other hyperparameter values in the Appendix
to ensure a fair comparison but found no substantial improvements. SCALE was not designed to include pretraining, and so we also experimented with a version that learns from T0 in an online streaming manner, but its performance was poor relative to PCMC.

\paragraph{STAM \cite{smith2021unsupervised}}:
We use the \href{https://github.com/CameronTaylorFL/stam}{original STAM code} with minor changes for computational efficiency. STAM was not designed to include a pretraining phase, and so the T0 data is presented as an additional streaming task, allowing STAM to use a portion of that data for initialization. We use similar hyperparameters to those in the original STAM paper for CIFAR10~\cite{krizhevsky2009learning}, but scaled up for larger resolution images. The full set of hyperparameters is in the Appendix.
\begin{figure*}[t!]
    \centering
    \includegraphics[width=1.1\textwidth]{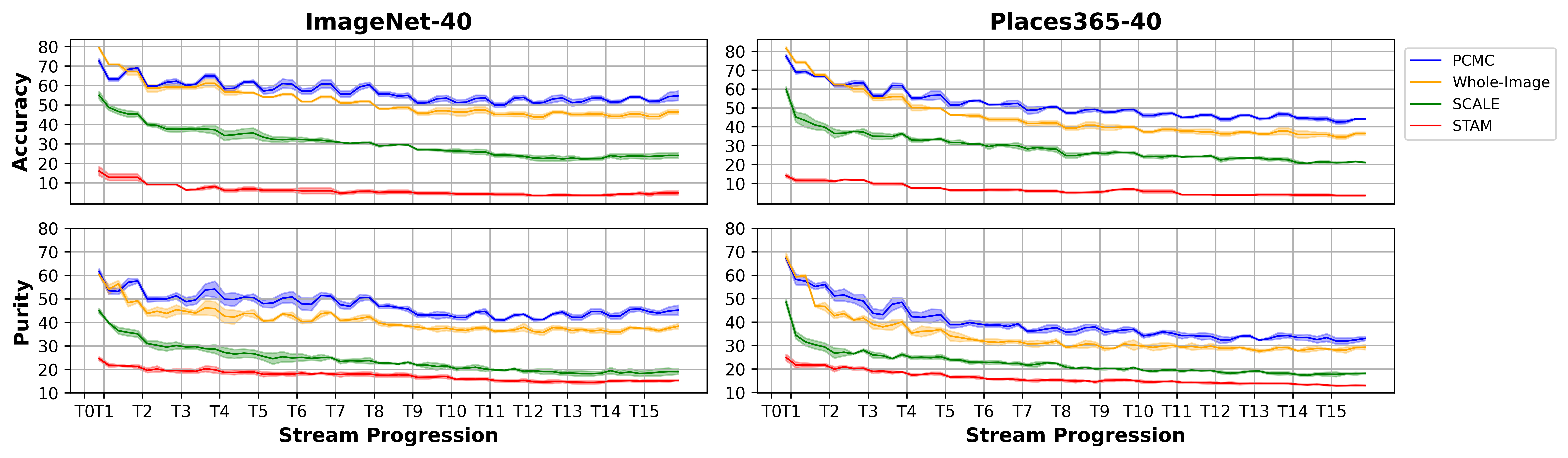}
    \caption{Classification and clustering performance comparisons between PCMC and baselines on the ImageNet-40 and Places365-40 streams. In both streams, the initial task T0 contains 10 classes, and each of the subsequent 15 tasks contains 2 classes each. Each task contains four evaluation points distributed evenly throughout the task, focusing on all classes seen so far. For the classification tasks, weuse 100 labeled examples per class and 100 test examples per class. We emphasize that these labeled examples are not used for representation learning during the stream -- they are only used to identify class-informative centroids. Average results over three independent seeded trials are shown, with error measured as $\pm$ one standard deviation.}
    \label{fig:baseline-comparisons}
\end{figure*}

\section{Experimental Results}

\subsection{Comparisons with Baselines}
Figure~\ref{fig:baseline-comparisons} shows classification and clustering performance of PCMC compared to several baselines. For ImageNet-40, the results are shown in the left column, with the first row representing classification accuracy and the second row showing clustering purity. Similarly, the Places365-40 results are shown in the right column. The performance of PCMC does not degrade significantly over time as new classes are added, highlighting its ability to incorporate new knowledge without major catastrophic forgetting. In the Appendix
, we also examine the performance of PCMC broken down by ``novel'' and ``past'' classes within each task. This demonstrates how the model quickly improves its understanding of new classes, with a significant boost after each sleep phase. Additionally, we examine forgetting over time, both pre- and post-sleep, showing how the sleep cycle helps the model disentangle class representations and improve performance on both past and novel classes compared to early evaluations during each task.

\subsection{Impact of $M$ on Performance}
\label{sec:mem-perf}
Table~\ref{tab:mem-compare} presents an evaluation of PCMC with three different values for $M$, along with the
\begin{wraptable}[17]{l}{0.53\textwidth}
\centering
\renewcommand{\arraystretch}{1.3}
\caption{Comparison of classification performance versus memory usage for PCMC and PCMC-NC without memory consolidation. Results are calculated over three independent seeded trials. $|LTM|$ is the maximum memory (in terms of images) utilized across all three trials, and accuracy is the average across all evaluations $\pm$ one standard deviation.}
\label{tab:mem-compare}
\scriptsize
\begin{tabular}[h!]{c|c|c|c|c|c}
    \hline
                    &      & \multicolumn{2}{|c|}{ImageNet-40} & \multicolumn{2}{|c}{Places365-40} \\
    \hline
                    &   M  & $|LTM|$         & Accuracy         & $|LTM|$                           & Accuracy                          \\
    \hline
    PCMC            & $10$ & $4647$  & $54.7 \pm 0.51$  & $4383$    & $51.6 \pm 0.63$ \\ 
                    & $20$ & $8819$  & $59.7 \pm 1.34$  & $8660$    & $55.7 \pm 0.98$ \\ 
                    & $30$ & $13250$ & $59.9 \pm 0.95$ & $14618$   & $55.5 \pm 0.88$ \\ 
    \hline 
    PCMC-NC         & $10$ & $6966$  & $56.3 \pm 0.84$   & $6040$    & $52.3 \pm 0.40$ \\
                    & $20$ & $12715$ & $60.0 \pm 1.21$  & $12800$   & $55.8 \pm 0.97$ \\
                    & $30$ & $18287$ & $60.0 \pm 0.54$  & $18983$   & $56.3 \pm 0.64$ \\
    \hline
\end{tabular}
\end{wraptable}
total amount of memory used at the end of the stream (in terms of whole images) and the average overall performance. The first three rows represent the actual PCMC algorithm, including the memory consolidation step, while the final three rows represent a PCMC ablation, without the memory consolidation step (PCMC-NC). For each value of $M$, the consolidation algorithm saves approximately 30\% of memory with minimal impact on overall performance. For this experiment, we consider a forgetting factor $k=2$, a minimum centroid capacity of $M_{min} = 5$, and $M_{init} = M$. In the Appendix
, we explore more values of $k$ and $M_{min}$, noting that more aggressive example pruning has a negative impact on performance.

\subsection{Novel vs Past Class Performance}
\label{res:novel-past}
\begin{figure*}[t!]
    \centering
    \begin{subfigure}[b]{0.49\textwidth}
        \includegraphics[width=\textwidth]{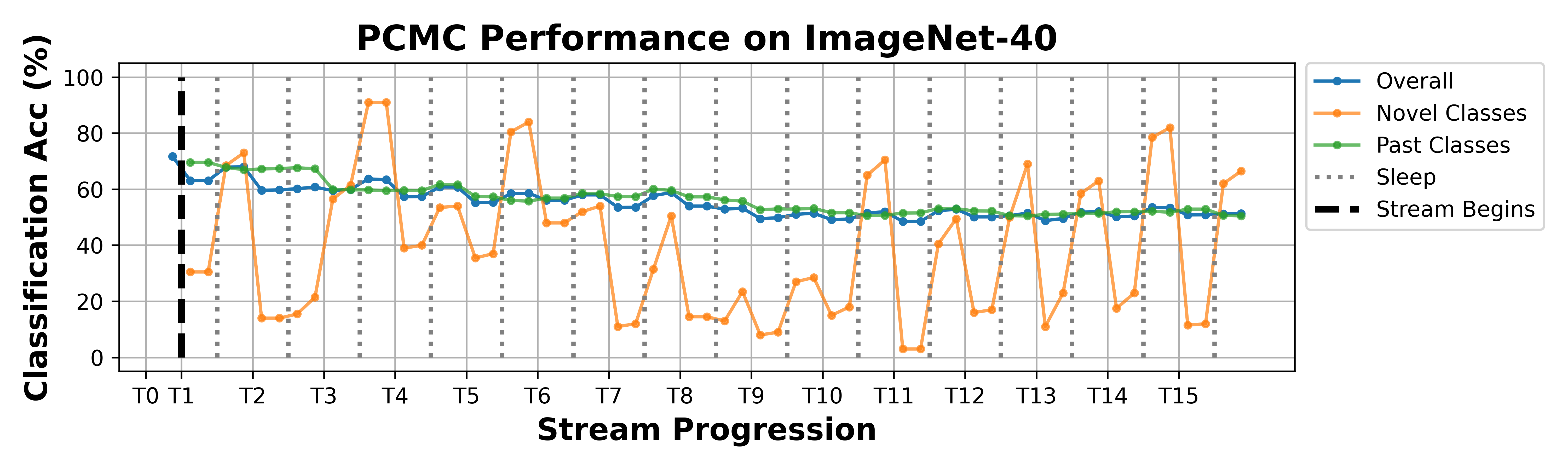}
        \caption{ImageNet-40}
        \label{fig:novel-im40}
    \end{subfigure}
    \hfill 
    \begin{subfigure}[b]{0.49\textwidth}
        \includegraphics[width=\textwidth]{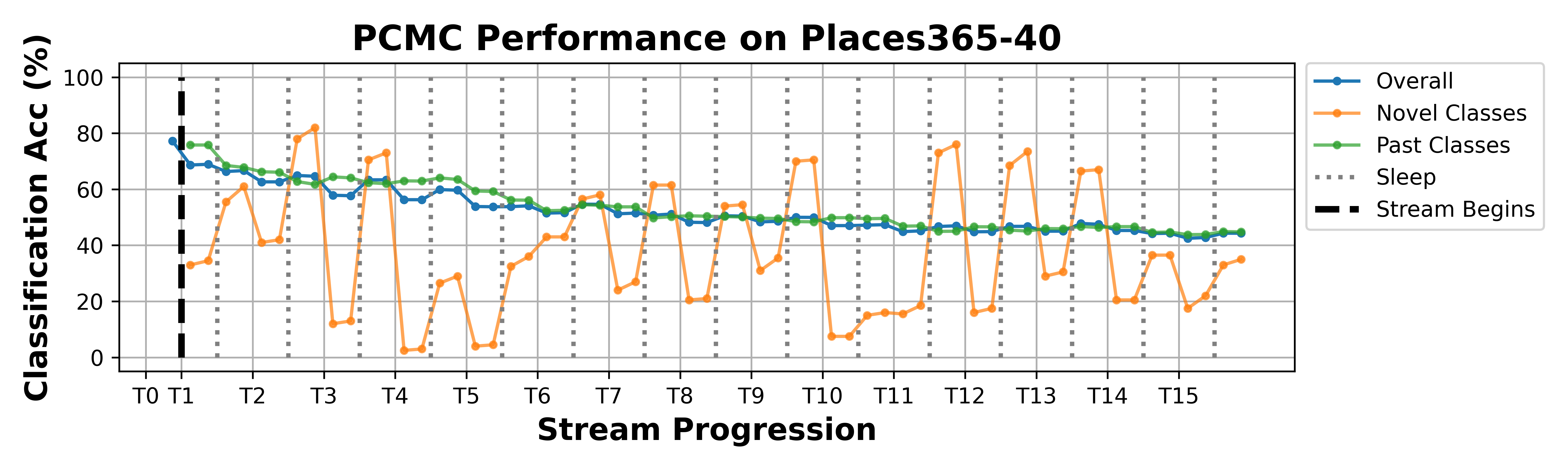}
        \caption{Places365-40}
        \label{fig:novel-pls}
    \end{subfigure}
    \caption{PCMC classification performance breakdown for a specific trial on the ImageNet-40 and Places365-40 streams. The orange curve represents the performance on the novel classes, the green curve represents performance on the past (previously observed) classes, and the blue curve represents the overall performance. The vertical grey dashed lines represent sleep cycles.}
    \label{fig:cornerstone-main}
\end{figure*}

Figure~\ref{fig:cornerstone-main} shows the performance of PCMC on ImageNet-40 and Places365-40, broken down into ``novel'' classes, ``past'' classes, and ``overall'', similar to Figure~\ref{fig:cornerstone}. This plot helps to understand how PCMC adapts to new classes while maintaining performance on previously seen classes.

We observe that in both figures, the model improves its performance on novel classes throughout the task. The extent of improvement varies depending on the difficulty of the new classes. If the model struggles to adapt to new classes, the performance on past classes drops when we enter a new task, as the previously ``novel'' classes now become ``past'' classes. However, if the model adapts well, the overall performance can increase slightly, maintaining consistent performance throughout the stream.

\subsection{Ablations}

\begin{figure*}[t!]
    \centering
    \begin{subfigure}[b]{0.32\textwidth}
        \includegraphics[width=\textwidth]{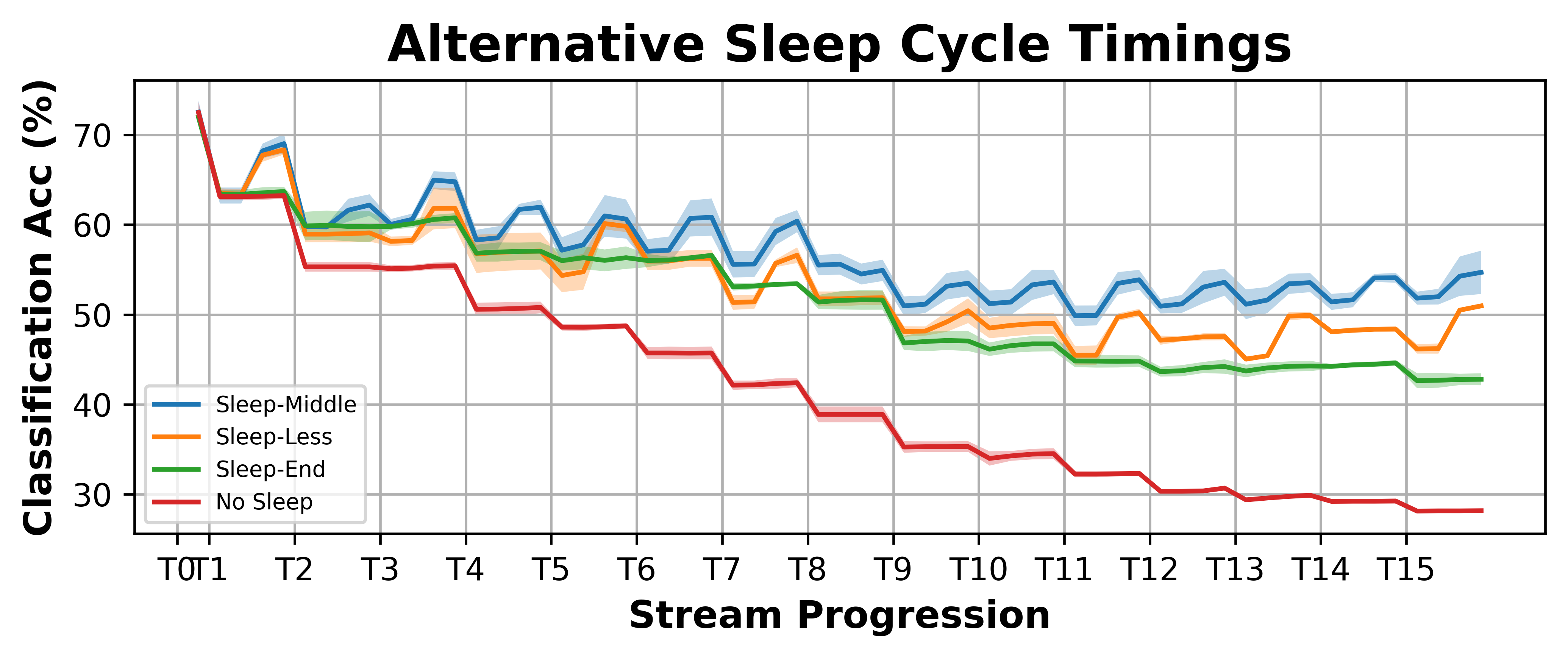}
        \caption{Sleep Cycle Ablation}
        \label{fig:sleep-timing-compare}
    \end{subfigure}
    \hfill 
    \begin{subfigure}[b]{0.32\textwidth}
        \includegraphics[width=\textwidth]{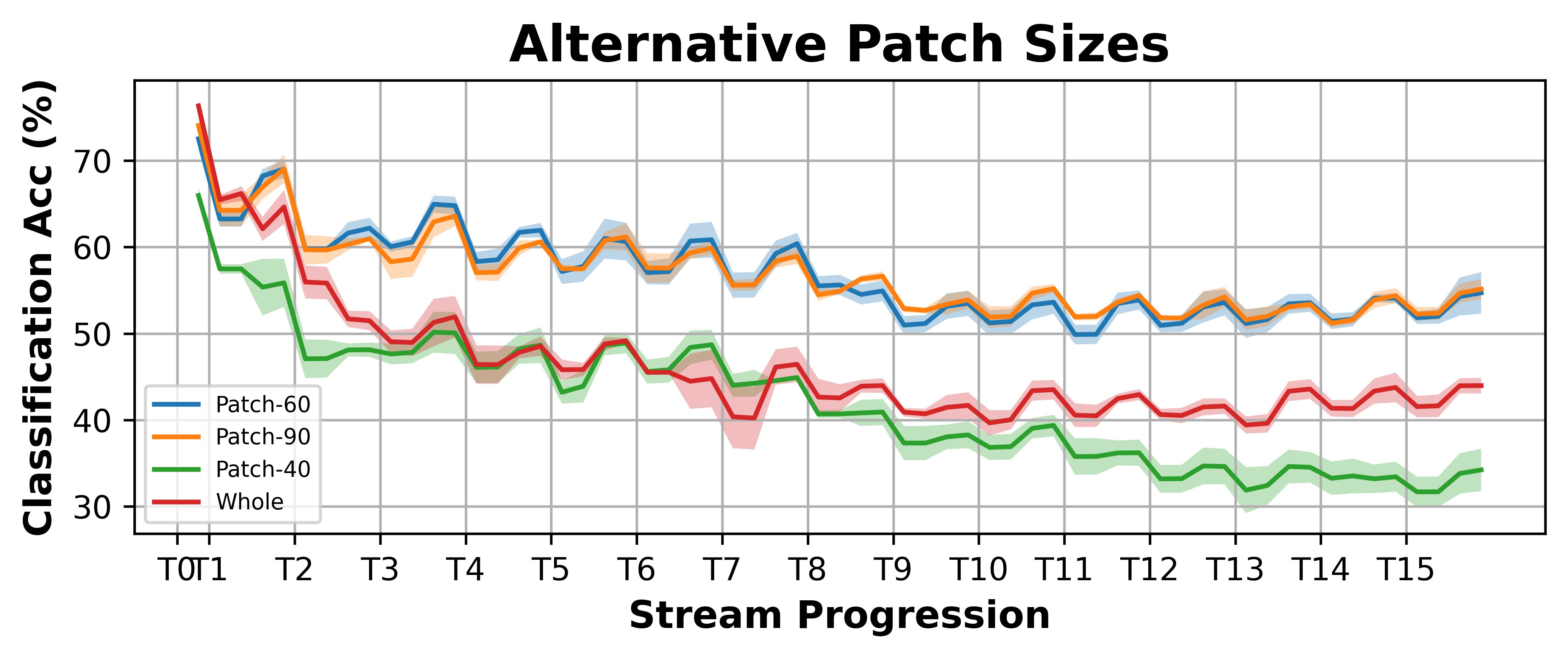}
        \caption{Patch Size Ablation}
        \label{fig:patch-size-compare}
    \end{subfigure}
    \hfill
    \begin{subfigure}[b]{0.32\textwidth}
        \includegraphics[width=\textwidth]{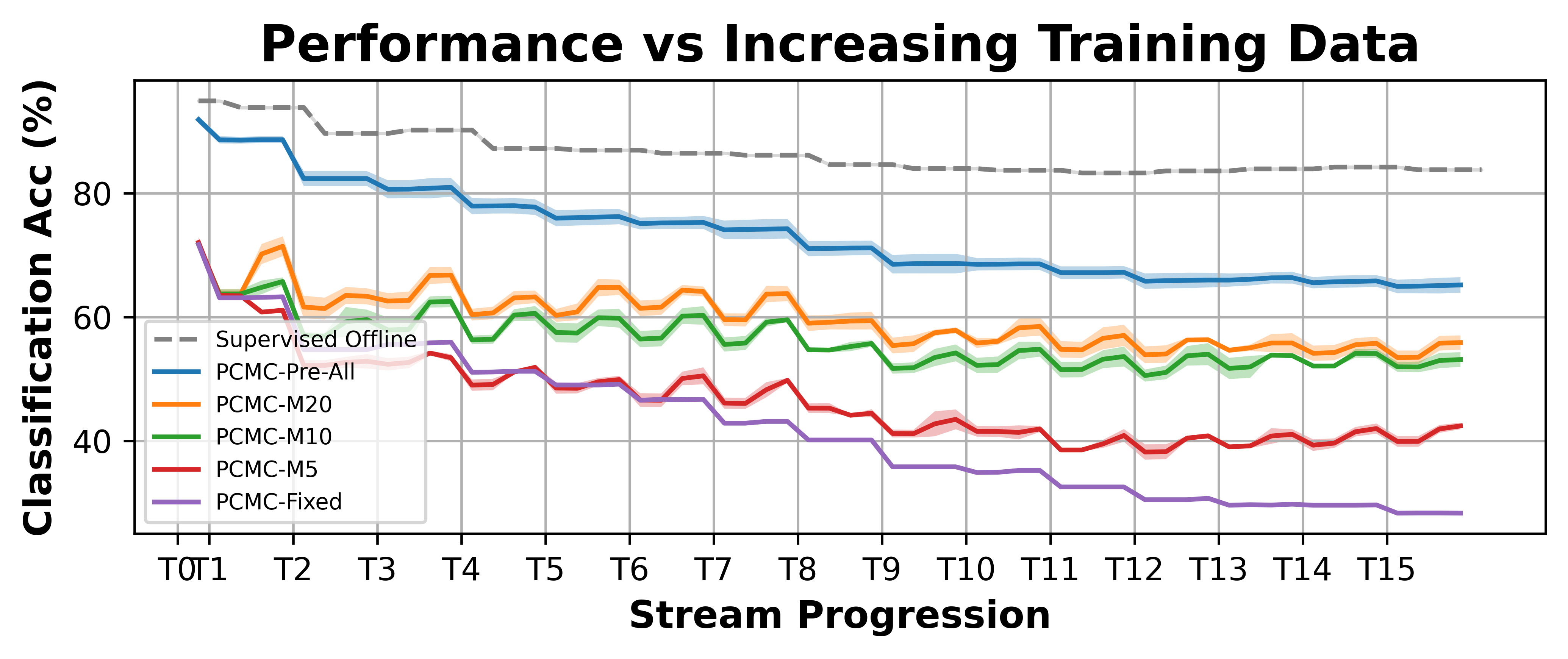}
        \caption{Lower \& Upper Bounds}
        \label{fig:pretrained-compare}
    \end{subfigure}
    \caption{PCMC ablations exploring the various modeling choices and performance bounds. a) explores the impact of the timing of the sleep cycle on the overall performance, b) examines the impact of patch size and also potential benefits of using patches at all, while c) compares PCMC with an upper and lower bound in terms of the quality of the encoder. Each experiment is the average over three independent seeded trials with error measured as $\pm$ one standard deviation.}
    \label{fig:ablations}
\end{figure*}

\paragraph{Sleep Timing:}
\label{res:sleep-timing}
Figure~\ref{fig:sleep-timing-compare} compares the classification performance of PCMC with different sleep cycle timings. The ``sleep-middle'' curve represents the traditional PCMC, where the model goes to sleep at fixed intervals in the middle of each task. The ``sleep-less'' curve represents the model sleeping in the middle of every other task, and the ``sleep-end'' curve represents the model sleeping at the end of each task. The most significant difference is observed in the ``no-sleep'' version of PCMC, while the ``sleep-end'' and ``sleep-less'' versions are closer to the primary PCMC approach. Interestingly, sleeping less is more beneficial than sleeping at the end of the task. We assume this is because the mid-task sleep cycle allows the model to integrate some understanding of the new distribution into the encoder, which can then be used to learn better centroids in the later half of the task. Future work could explore a version of PCMC that detects shifts in the data distribution (possibly via its existing novelty detection mechanisms) and goes to sleep after collecting a sufficient number of novel examples.

\vspace{-1em}
\paragraph{Patch Size Comparison:}
\label{res:patch-compare}
Figure~\ref{fig:patch-size-compare} shows an ablation study comparing PCMC with different patch sizes. The key takeaway is that patch sizes that are either too small or too large significantly harm performance. However, several intermediate patch sizes yield similar performance. In our primary experiments, we chose a patch size of 60 over 90 because it allows us to save about 30\% of the total memory budget. Even with increased memory allowance, the smaller patch size of 40 and the whole-image version of PCMC do not compare. This comparison highlights the benefits of patches both in terms of memory efficiency and in incorporating new data into the model's learned representations.

\vspace{-1em}
\paragraph{Comparison with Upper \& Lower Bounds:}
Figure~\ref{fig:pretrained-compare} demonstrates the performance of PCMC with a frozen encoder that is pretrained in an offline fashion with all classes from the entire stream (PCMC-Pre-All), a frozen encoder that is pretrained with self-supervision only on the T0 data (PCMC-Fixed), and a series of PCMC models with dynamic encoders that use progressively more memory for storing raw examples. We also include a ``Supervised Offline'' upper bound that is based on a ResNet18 encoder with the pretrained ImageNet-1k weights and a k-NN classifier. The Supervised Offline and PCMC-Pre-All versions are meant to act as upper bounds, when the models operate with ``ideal'' encoders. Note that as the number of classes increases, even this ideal encoder-based model suffers from a decrease in performance due to increased task complexity. This may seem surprising given that the pretrained encoder in this case was trained on the entire ImageNet-1k dataset, with labels. However, the use of PCMC's classification algorithm and the limited number of labeled examples per class make this task more challenging than typical 40-way classification. As we increase the amount of memory that PCMC utilizes, we see that the gap in overall performance closes and it is much closer to the Pre-All upper-bound. However, the gap with the supervised offline approach is still large.

\paragraph{Alternative Encoder Architectures:}
We also compare PCMC, SCALE, and the Whole-Image baseline with various encoder sizes. Table~\ref{tab:arch-compare} showcases the performance of the three approaches utilizing neural network-based encoders. The performance is shown in terms of the average classification accuracy and clustering purity throughout all evaluations in the stream. The three architectures considered are ResNet18, ResNet34, and ResNet50, with their parameter counts shown in the first column of the table. When comparing approaches that utilize the the same backbone, PCMC's performance is the best. Additionally, for both classification and clustering, PCMC outperforms all versions of the Whole-Image baseline and SCALE with just the ResNet18 backbone. With SCALE, we see a decrease in performance as the encoder size increases, likely due to the limited number of training iterations in SCALE. This makes learning more difficult with a larger number of parameters and causes performance to drop throughout the stream.

\section{Related Work}
\label{sec:related}

{\bf Online Unsupervised Continual Learning: }
Variations of the O-UCL problem have been studied under different names. \citep{smith2021unsupervised} was the first to introduce a  similar problem called ``Unsupervised Progressive Learning'' (UPL). While O-UCL and  UPL  share some similar characteristics, there are also important differences. UPL does not include a pretraining phase with some initial data. Also, UPL does not allow the model to save any raw examples (images or patches) during the stream. In the case of O-UCL, we allow for some data to be available upfront for pretraining, and include the corresponding classes in the evaluation set. 
\begin{wraptable}[20]{r}{0.53\textwidth}
\centering
\renewcommand{\arraystretch}{1.2}
\caption{Comparison of PCMC, SCALE, and Whole-Image Baseline on ImageNet-40 with several different backbone architectures. Results show average accuracy and purity across all evaluations throughout the stream, consisting of three independently seeded trials showing mean $\pm$ one standard deviation.}
\label{tab:arch-compare}
\footnotesize
\begin{tabular}[h!]{c|c|c|c}
    \hline
                    &  Params       & Accuracy          & Purity \\
    \hline
    PCMC-ResNet18   & $11.7$ M      & $56.6 \pm 1.3$    & $47.2 \pm 0.2$ \\ 
    PCMC-ResNet34   & $21.8$ M      & $57.9 \pm 1.6$    & $47.6 \pm 1.0$ \\ 
    PCMC-ResNet50   & $25.6$ M      & $\bf 58.6 \pm 0.8$    & $\bf 48.3 \pm 0.3$ \\              
    \hline 
    SCALE-ResNet18   & $11.7$ M      & $30.4 \pm 0.8$    & $24.1 \pm 0.7$ \\ 
    SCALE-ResNet34   & $21.8$ M      & $16.3 \pm 0.6$    & $18.0 \pm 0.5$ \\ 
    SCALE-ResNet50   & $25.6$ M      & $22.9 \pm 1.3$    & $20.6 \pm 0.5$ \\             
    \hline
    Whole-ResNet18   & $11.7$ M      & $51.6 \pm 0.6$    & $40.9 \pm 0.4$ \\ 
    Whole-ResNet34   & $21.8$ M      & $53.9 \pm 0.3$    & $43.3 \pm 0.3$ \\ 
    Whole-ResNet50   & $25.6$ M      & $55.3 \pm 0.5$    & $44.9 \pm 1.4$ \\             
    \hline

\end{tabular}
\end{wraptable}
SCALE \citep{yu2022scale} and DAA \citep{michel2023domain}  address a problem that is similar to O-UCL but they allow for storage of entire images and do not include an initialization task. SCALE combines mixture replay with contrastive learning to learn from an unlabeled stream with a changing distribution, while maintaining a fixed memory of examples to avoid forgetting. DAA approaches the problem similarly to SCALE but focuses on novel augmentation strategies applied to stream and replay images. Specifically, DAA utilizes MixUp, CutMix, and Style transfer to combine stored examples with new examples. Unfortunately, we were not able to reproduce the DAA method and its original implementation is not yet publicly available. OUPN \citep{ren2021online} studies another similar setting but focuses on understanding the current distribution rather than avoiding catastrophic forgetting over time. It learns an online mixture of Gaussians similar to \citep{rao2019continual} but instead of just using a reconstruction loss, the model contrasts cluster (mixture component) assignments. 

{\bf Continual Learning: }
Continual learning learns a sequence of  different tasks over time. Typically, it is assumed that each task has  a fixed data distribution~\citep{hsu2018re}. Some approaches focus on regularizing the ``important'' weights learned in previous tasks~\citep{kirkpatrick2017overcoming, li2017learning, rannen2017encoder, aljundi2018memory, aljundi2019task, lopez2017gradient}. Others focus on storing representative examples or representations and replaying them during future tasks~\citep{aljundi2019gradient, buzzega2020dark, rebuffi2017icarl, rolnick2019experience, shin2017continual}. More recently, some works have focused on improving the efficiency of replay memories~\citep{brignac2023improving, hurtado2023memory, bai2023saliency}. Some methods focus on the supervised online continual learning setting~\citep{lee2023online, hayes2020remind, hayes2020lifelong}, while others focus on memory or compute-constrained environments~\citep{ghunaim2023real, demosthenous2021continual, hayes2022online, fini2020online, harun2023siesta}. These works may have slightly different learning criteria and objectives but they all differ substantially from our work because  they rely on labels. 

Our work does not assume knowledge of task boundaries, which is similar to ``Task-Free Continual Learning'' (TFCL)~\citep{aljundi2019task} but that work does not consider learning from a stream. 

The streaming aspect of O-UCL is also similar to ``Online Continual Learning'' (OCL)~\citep{aljundi2019gradient} but the data in O-UCL is unlabeled. Several other works under the umbrella of continual learning have looked at similar problems but they do not consider all challenges at the same time. For instance, some works examine the unsupervised continual learning problem with a focus on learning generative models in a continual but offline manner~\citep{rao2019continual, ye2020learning, lee2020neural, ramapuram2020lifelong, achille2018life}. Other works focus on self-supervision for good representations or use a contrastive loss to contrast between new and old samples across tasks~\citep{morawiecki2022hebbian, ni2021self, zhang2020self, cha2021co2l, li2022continual, madaan2021representational, fini2022self, lin2022continual}. 

{\bf Self-Supervised Learning: }
Self-supervised learning focuses on learning to extract useful representations from unlabeled data. Some of the earlier computer vision self-supervised learning methods involved utilizing pretext tasks such as inpainting~\citep{pathak2016context}, colorization~\citep{deshpande2015learning, larsson2016learning, zhang2016colorful}, denoising~\citep{vincent2008extracting}, and making predictions about rotation or relative position~\citep{gidaris2018unsupervised}. More recently, instance-based approaches that contrast augmented versions of different images have been proposed~\citep{chen2020improved, chen2020simple, chen2021exploring, he2020momentum, grill2020bootstrap}. Expanding on this idea, cluster-based approaches that contrast cluster assignments of augmented versions of different images have become much more popular~\citep{li2020prototypical, li2022neural, caron2018deep, caron2020unsupervised}. Our work is designed to utilize an instance-based contrastive objective for training the encoder, but the O-UCL framework is agnostic to the specific method.

\vspace{-0.5em}
\section{Conclusions}
This work makes two contributions. First, we present the Online Unsupervised Continual Learning problem and make the case that it captures many real-world applications, even though it is also a much harder problem to define mathematically and solve in practice. The second contribution is a method (PCMC) that attempts to solve the O-UCL problem.  

PCMC is designed to operate effectively in complex, real-world environments, learning from non-stationary and single-pass data streams without the need for external supervision or predefined knowledge. PCMC's dynamic encoder and patch-based contrastive learning allow for more nuanced and adaptable feature extraction. This is complemented by an innovative sleep-cycle mechanism for periodic encoder retraining, ensuring continuous adaptation. Using two streams of natural images, we show that PCMC outperforms all baselines in both classification and clustering evaluation tasks. Additionally, ablation experiments explore the design choices behind PCMC, and showcase its memory utilization. 

The experimental evaluation also shows that PCMC is not a perfect solution, and that it does not manage to completely avoid catastrophic forgetting. Consequently, we expect and hope that significant improvements will be feasible in the future. Follow-up work can also explore how to avoid the need for storing  raw patches, potentially saving only patch embeddings in memory, and utilizing a decoder to restore the raw patches. Another improvement can focus on the sleep cycle, so that the learner sleeps when it needs to retrain its encoder and not periodically.  

\paragraph{Acknowledgements}
This work was supported by the Lifelong Learning Machines
(L2M) program of DARPA/MTO: Cooperative Agreement
HR0011-18-2-0019.

\bibliography{collas2024_conference}
\bibliographystyle{collas2024_conference}

\end{document}